\pdfoutput=1
\documentclass[11pt]{article}

\usepackage{EMNLP2023}

\usepackage{times}
\usepackage{latexsym}

\usepackage[T1]{fontenc}
\usepackage[utf8]{inputenc}

\usepackage{microtype}

\usepackage[shortlabels]{enumitem}
\usepackage{inconsolata}
\usepackage{booktabs}
\usepackage{graphicx}
\usepackage{multirow}
\usepackage{amsmath}
\usepackage{amssymb}
\usepackage{euflag}
  {\list{}{\leftmargin=#1\rightmargin=#1}\item[]}%
  {\endlist}
\title{On the Interplay between Fairness and Explainability}

\newcommand\ie{\emph{i.e.}}
\newcommand\eg{\emph{e.g.}}
\newcommand\biased{BIOS$_\texttt{biased}$}
\newcommand\balanced{BIOS$_\texttt{balanced}$}

\author{Stephanie Brandl \quad Emanuele Bugliarello \quad Ilias Chalkidis\\
Department of Computer Science, University of Copenhagen, Denmark\\
\texttt{\{brandl, emanuele, ilias.chalkidis\}@di.ku.dk}}

\begin{document}
\maketitle
\begin{abstract}
In order to build reliable and trustworthy NLP applications, models need to be both fair across different demographics and explainable. Usually these two objectives, \emph{fairness} and \emph{explainability}, are optimized and/or examined independently of each other. Instead, we argue that forthcoming, trustworthy NLP systems should consider both.
In this work, we perform a first study to understand how they influence each other: do \emph{fair(er)} models rely on \emph{more plausible} explanations? and vice versa. To this end, we conduct experiments on two English multi-class text classification datasets, BIOS and ECtHR, that provide information on gender and nationality, respectively, as well as human-annotated rationales. 
We fine-tune pre-trained language models with several methods for (i) bias mitigation, which aims to improve fairness; (ii) rationale extraction, which aims to produce plausible explanations.
We find that bias mitigation algorithms do not always lead to fairer models. Moreover, we discover that empirical fairness and explainability are orthogonal.
\end{abstract}

\section{Introduction}
Fairness and explainability are crucial factors when building trustworthy NLP applications. This is true in general, but even more so in critical and sensitive applications such as medical~\cite{gu-etal-2020-blurb} and legal~\cite{chalkidis-etal-2022-lexglue} domains, as well as in algorithmic hiring processes \cite{schumann2020we}. AI trustworthiness and governance are no longer wishful thinking since more and more legislatures introduce related regulations for the assessment of AI technologies, such as the EU \citet{eu-ai-act}, the US \citet{us-ai-act}, and the Chinese \citet{china-ai-act}. Therefore, it is important to ask and answer challenging questions that can lead to safe and trustworthy AI systems, such as how fairness and explainability interplay when optimizing for either or both.

\begin{figure}[t]
    \centering
    \resizebox{\columnwidth}{!}{
    \includegraphics{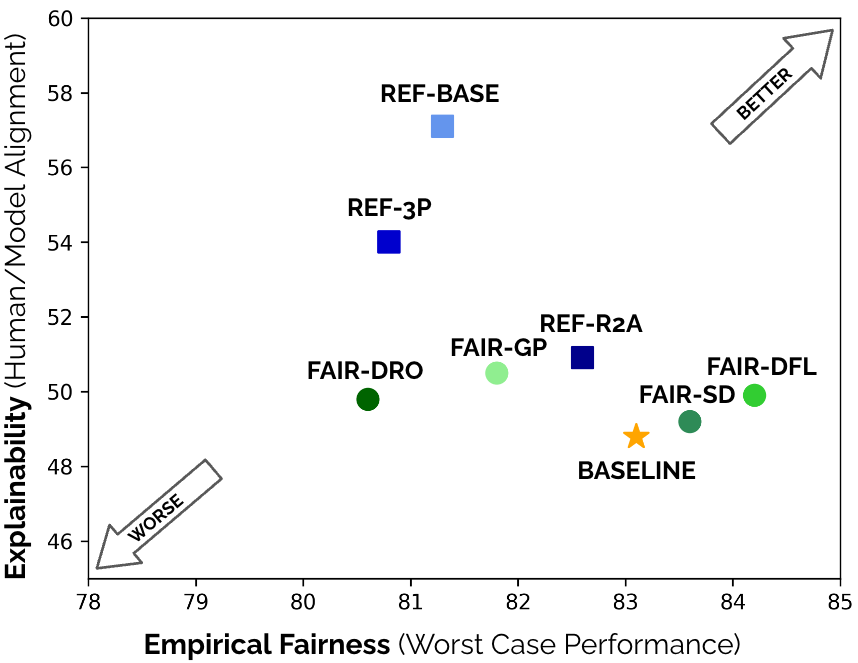}
    }
    % \vspace{-1mm}
    \caption{Interplay between \emph{empirical fairness}, measured via worst-case performance, and \emph{explainability} measured via human/model alignment, of different methods (Section~\ref{sec:methods}) optimizing for fairness (\textsc{fair}), explainability (\textsc{ref}), or none (\textsc{baseline}) on the ECtHR dataset. All methods, including the baseline, are built upon fine-tuned RoBERTa models. The results here suggest that the two dimensions are independent.}
    \label{fig:enter-label}
    \vspace{-4mm}
\end{figure}

So far in the NLP literature, model explanations\footnote{We refer to both the feature attribution scores assigned by models (binary and continuous) and the binary annotations by humans as \emph{rationales} throughout the paper, and also use the term \emph{(model) explanations} for the former.} are used to detect and mitigate how fair or biased a model is \cite{balkir-etal-2022-challenges} or to assess a user's perception of a model's fairness \cite{zhou2022towards}. Those are important use cases of explainability but we argue that we should further aim for improving one when optimizing for the other to promote trustworthiness holistically across both dimensions. 
% Our work is based on the following hypothesis: 
% \begin{myquote}{0.05in}
% \emph{Models providing factual rationales should, in theory, not rely on spurious correlations, and thus should also be fair(er). Likewise, models that are fair across different demographics, should also provide factual rationales.\footnote{By \underline{factual rationales}, we refer to plausible rationales that have a causal relationship to the expected outcome (prediction) in a given task, and by \underline{fair models}, we refer to models that perform approximately equally across groups and/or improve performance for the worst-performing group.}}
% \end{myquote}

To analyze the interplay between fairness and explainability, we optimize neural classifiers for one or the other during fine-tuning, and then evaluate both afterwards (Figure~\ref{fig:enter-label}). 
We do so across two English multi-class classification datasets.
First, we analyze a subset of the BIOS dataset~\cite{dearteaga-etal-2019-bios}. This dataset contains short biographies for occupation classification. We consider a subset of 5 medical professions that also includes human annotations on 100 biographies across this subset \cite{eberle-etal-2023-contrast-bios}. We evaluate model-based rationales extracted via (i) LRP \cite{ali-etal-2022-lrp} or (ii) rationale extraction frameworks (REFs;~\citealt{lei-etal-2016-rationalizing}), while also examining fairness with respect to gender. 
Second, we also conduct similar experiments with the ECtHR dataset~\cite{chalkidis-etal-2021-paragraph} for legal judgment forecasting on cases from the European Court of Human Rights (ECHR), both to evaluate paragraph-level rationales and to study fairness with respect to the nationality of the defendant state. 
\vspace{-1mm}
\paragraph{Contributions.}
Our main contributions in this work are the following:
\textbf{(i)} We examine the \emph{interplay} between two crucial dimensions of trustworthiness: \emph{fairness} and \emph{explainability}, by comparing models that were fine-tuned using five fairness-promoting techniques (Section~\ref{sec:opt-fair}) and three rationale extraction frameworks (Section~\ref{sec:opt-rationale}) on two English multi-class classification dataset (BIOS and ECtHR).
\textbf{(ii)} Our experiments on both datasets (a) confirm recent findings on the independence of bias mitigation and empirical fairness~\cite{cabello2023independence}, and (b) show that also empirical fairness and explainability are independent.

\section{Related Work}
\vspace{-2mm}
\paragraph{Bias mitigation /  fairness.}
The literature on inducing fairer models from biased data is rapidly growing (see \citealt{mehrabi2021survey,maklouf2021,ding2021retiring} for recent surveys). Fairness is often conflated with bias mitigation, although they have been shown to be orthogonal: reducing bias, such as representational bias, may not lead to a fairer model in terms of downstream task performance~\cite{cabello2023independence}. In this work, we follow the definition of \citet{shen-etal-2022-representational} and target \emph{empirical fairness} (performance parity) that may not align with \emph{representational fairness} (data parity). This means that we adopt a capability-centered approach to fairness and define fairness in terms of performance parity \cite{pmlr-v80-hashimoto18a} or equal risk \cite{donini18empirical}. The fairness-promoting learning algorithms that we evaluate are discussed in detail in Section~\ref{sec:methods}.
\vspace{-1mm}
\paragraph{Explainable AI (XAI) for fairness.}
% explanations to detect fairness
Explanations have been used to improve user's perception and judgement of fairness \cite{shulner2022fairness, zhou2022towards}. 
\citet{balkir-etal-2022-challenges} give an overview of the *ACL literature where explainability is applied to detect and mitigate bias.
They predominantly find work on uncovering and investigating bias for hate speech detection. 
% square one
Recently, also \citet{ruder-etal-2022-square} call for more multi-dimensional NLP research where fairness, interpretability, multilinguality and efficiency are combined.
The authors only find work by \citet{vig2020investigating} who use explainability to find specific parts of a model that are causally implicated in its behaviour. With this work, we want to extend this line of research from `XAI for fairness' to `XAI and Fairness'.
\vspace{-1mm}
\paragraph{Post-hoc XAI.}
XAI methods commonly rely on saliency maps extracted post-hoc from a model using attention scores~\cite{bahdanau2015nmtranslation,abnar-zuidema-2020-quantifying}, gradients~\cite{voita-etal-2019-analyzing,wallace-etal-2019-allennlp,ali-etal-2022-lrp}, or perturbations \cite{RibeiroSG16,alvarez2017,Murdoch2018} at inference time. These can be seen as an approximation of identifying which features (tokens) the model relied on to solve a given task for a specific example. Such methods do not necessarily lead to \emph{faithful} explanations~\cite{jacovi-goldberg-2020-towards}. Following \citet{deyoung-etal-2020-eraser}, faithfulness can be defined as the combination of \emph{sufficiency}---tokens with the highest scores correspond to a sufficient selection to reliably predict the correct label---\emph{and} \emph{comprehensiveness}---all indicative tokens get attributed relatively high scores.
\vspace{-1mm}
\paragraph{Rationale extraction by design.} 
Unlike post-hoc explanations, \emph{rationale extraction} frameworks \emph{optimize} for rationales that support a given classification task and are faithful by design, \ie, predictions are based on selected rationales by definition.

\citet{lei-etal-2016-rationalizing} were the first to propose a framework to produce short coherent rationales that could replace the original full texts, while maintaining the model's predictive performance. 
The rationales are extracted by generating binary masks indicating which words should be selected; and two additional loss regularizers were introduced, which penalize long rationales and sparse masks that would select non-consecutive words. 

Recently, several studies have proposed improved frameworks that rely mainly on complementary adversarial settings that aim to produce better (causal, complete) rationales and close the performance gap compared to models using the full input~\cite{yu-etal-2019-rethinking,chang-etal-2019-game,jain-etal-2020-learning,yu2021understanding}. The rationale extraction frameworks we evaluate are detailed in Section~\ref{sec:methods}.

\paragraph{XAI \emph{and} fairness.}
\citet{mathew2021hatexplain} release a benchmark for hate speech detection where human annotations are used as input to the model to improve performance and fairness across demographics. They evaluate both faithfulness of post-hoc explanations as well as performance disparity across communities affected by hate speech.
\citet{he-etal-2022-controlling} propose a new debiasing framework that consists of two steps.
First, they apply the rationale extraction framework (REF) from \citet{lei-etal-2016-rationalizing} to detect tokens indicative of a given \textit{bias} label, \eg, gender. In a second step, those rationales are used to minimize bias in the task prediction. 
% Rationales from the REF are evaluated based on faithfulness. 
% Both of these proposed approaches are valuable and important in the process of achieving fair \emph{and} explainable models. 

With this work, we aim to complement prior work by systematically evaluating the impact of optimizing for fairness on explainability and vice versa, considering many different proposed techniques (Section~\ref{sec:methods}). Moreover, we consider both post-hoc explanations extracted from standard Transformer-based classifiers, as well as rationale extraction frameworks evaluating model-based explanations (rationales) in terms of faithfulness and alignment with human-annotated rationales.

\begin{figure*}[t]
    \centering
    \resizebox{\textwidth}{!}{
        \includegraphics{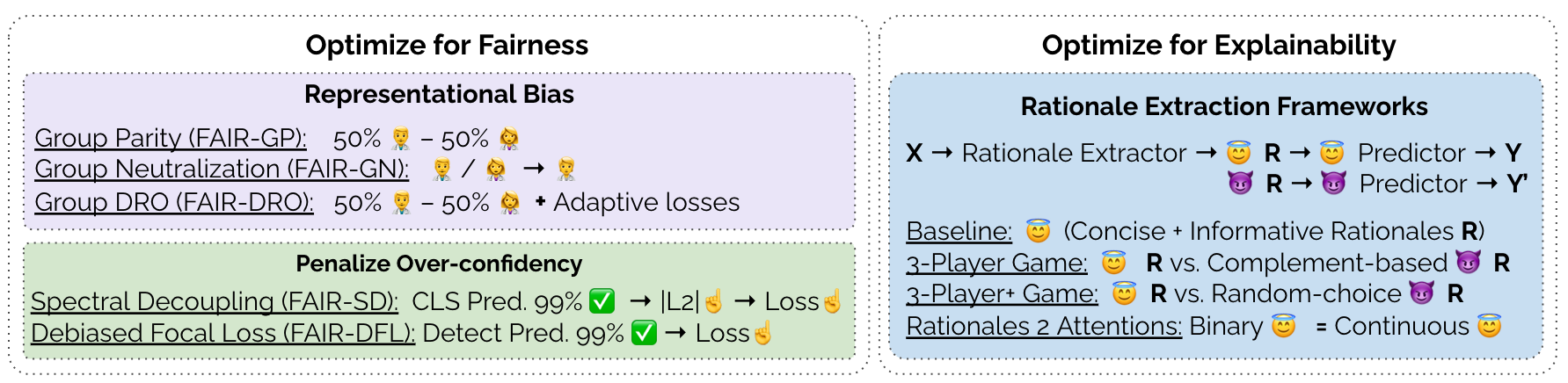}
    }
    \vspace{-6mm}
    \caption{A short description / depiction of the \emph{fairness-promoting} (Section~\ref{sec:opt-fair}) and \emph{explainability-promoting} (Section~\ref{sec:opt-rationale}) examined methods. The emojis represent male/female/neutral, and main, and adversarial modules.}
    \label{fig:methods-demo}
    \vspace{-5mm}
\end{figure*}

\section{Datasets}
\label{sec:datasets}

\paragraph{BIOS.} The BIOS dataset~\cite{dearteaga-etal-2019-bios} comprises biographies labeled with occupations and binary gender in English. This is an occupation classification task, where bias with respect to gender can be studied. In our work, we consider a subset of 10,000 (8K train / 1K validation / 1K test) biographies targeting 5 medical occupations (\emph{psychologist}, \emph{surgeon}, \emph{nurse}, \emph{dentist}, \emph{physician}) published by \citet{eberle-etal-2023-contrast-bios}. 
For these occupations, as shown in Table~\ref{tab:bios_stats}, there is a clear gender imbalance, \eg, 91\% of the nurses are female, while 85\% of the surgeons are male. 
We also compare with human rationales provided for a subset of 100 biographies.
% Here, each biography has been annotated by three annotators, and we consider the tokens that the \emph{majority} of annotators selected as the ground truth.

For control experiments on the effect of bias mitigation methods, we also create a synthetic extremely unbalanced (\emph{biased}) version of the train and validation split of BIOS, we call this version \biased.
Here, our aim is to amplify gender-based spurious correlations in the training subset by keeping only the biographies where all psychologists and nurses are female persons; and all surgeons, dentists, and physicians are male persons. 
Similarly, we create a synthetic balanced (\emph{debiased}) version of the dataset which we call \balanced.
Here, our objective is to mitigate gender-based spurious correlations by down-sampling the majority group per medical profession.
By doing so, in \balanced, both genders are equally represented per profession.

\begin{table}[t]
\centering
    \resizebox{0.85\columnwidth}{!}{
        \begin{tabular}{lrr}
    \toprule
    \multicolumn{3}{c}{\bf BIOS} \\
    \midrule
    Occupation & \multicolumn{1}{c}{Male} & \multicolumn{1}{c}{Female}\\
    \midrule
    Psychologist         &  822 (37\%) & 1378 (63\%) \\
    Surgeon              & 1090 (85\%) & 190 (15\%)\\
    Nurse                &  152 (09\%) & 1486 (91\%) \\
    Dentist              &  996 (65\%) & 537 (35\%) \\
    Physician            &  650 (48\%) & 699 (52\%)\\
    \\[-4mm]
    % \cmidrule{2-3}
    \emph{Total}                &  3710 (46\%) & 4290 (54\%) \\
    \midrule
    \multicolumn{3}{c}{\bf ECtHR} \\
    \midrule
    ECHR Article & \multicolumn{1}{c}{E. European} & \multicolumn{1}{c}{Rest} \\
    \midrule
    3 -- Proh. Torture  & 303 (88\%)  &   42 (12\%) \\
    5 -- Liberty        & 382 (88\%)  &   51 (12\%) \\
    6 -- Fair Trial     & 1776 (80\%) &  454 (20\%) \\
    8 -- Private Life   & 129 (55\%)  &  104 (45\%) \\
    P1.1 -- Property    & 228 (88\%)  &   31 (12\%) \\
    \\[-4mm]
    % \cmidrule{2-3}
    \emph{Total} & 2818 (80\%) & 682 (20\%) \\ \bottomrule
\end{tabular}}
    \caption{Label and demographic attribute distribution across the training sets of the BIOS and ECtHR datasets.}
    \label{tab:bios_stats}
    \vspace{-6mm}
\end{table}

% \vspace{-2mm}
\paragraph{ECtHR.} The ECtHR dataset \cite{chalkidis-etal-2021-paragraph} contains 11K cases from the European Court of Human Rights (ECHR) written in English. The Court hears allegations that a European state has breached human rights provisions of the European Convention of Human Rights (ECHR). For each case, the dataset provides a list of \emph{factual} paragraphs (facts) from the case description. Each case is mapped to \emph{articles} of the ECHR that were violated (if any). The dataset also provides silver (automatically extracted) paragraph-level rationales. We consider a subset of 4,500 (3.5K train / 500 validation / 500 test) single-labeled cases for five well-supported ECHR articles and the \emph{defendant state} attribute.
In practice, we use a binary categorization of the defendant states---Eastern European vs.~the Rest---to assess fairness, similar to \citet{chalkidis-etal-2022-fairlex}. Table~\ref{tab:bios_stats} shows a clear defendant state imbalance across most of the ECHR articles except for Article 8.
% For instance, 88\% of the cases are against E. European states for ECHR articles~3, 5, 6, and P1.1; while there is almost an equal number of cases (55/45\%) for Article~8.

\section{Methodology}
\label{sec:methods}
We fine-tune classifiers optimizing for either fairness (Section~\ref{sec:opt-fair}), explainability (Section~\ref{sec:opt-rationale}), or none, alongside the main classification task objective (Figure~\ref{fig:methods-demo}). The baseline classifier uses an $n$-way classification head on top of the Transformer-based text encoder~\citep{transformer}, and it is optimized using the cross-entropy loss against the gold labels~\cite{devlin-etal-2019-bert}.

\subsection{Optimizing for Fairness}
\label{sec:opt-fair}

We use a diverse set of 5 fairness-promoting algorithms that are connected to two different approaches: (a) mitigating \emph{representational bias} (\textsc{fair-gp}, \textsc{fair-gn}, \textsc{fair-dro}), and (b) penalizing \emph{over-confident predictions} (\textsc{fair-sd}, \textsc{fair-dfl}).

% \subsubsection*{Representational bias}
\paragraph{Representational bias}
\emph{Representational bias} (\eg, more data points for male vs. female surgeons) is considered a critical factor that may lead to performance disparity across demographic groups, as a model may rely on the protected attribute (\eg, gender) as an indicator for predicting the output class (\eg, occupation). We consider three methods to mitigate such effects:
\vspace{-2mm}
\newcounter{saveenum}
\begin{enumerate}[align=left,leftmargin=*,wide = 0pt,itemsep=-2pt, label={\roman*})]
\item \textbf{Group Parity} (\textsc{fair-gp}) where we over-sample the minority group examples per class up to the same level as the majority ones~\citep{Sun2009ClassificationOI}. For instance, by up-sampling biographies of male nurses and female surgeons in the BIOS dataset.
\item \textbf{Group Neutralization} (\textsc{fair-gn}), where we replace (normalize) attribute-related information.
For instance, for gender in BIOS, we replace gendered pronouns (\eg`he/him', 'she/her'), and titles (\eg `Mr', `Mrs'), with gender-neutral equivalents, such as `they/them' and `Mx' \cite{brandl-etal-2022-conservative},  while also replacing personal names with a placeholder name~\cite{hall-maudslay-etal-2019-name}, such as `Sarah Williams' with `Joe Doe'.
\item \textbf{Group Robust Optimization} (\textsc{fair-dro}) where we use GroupDRO as proposed by \citet{sagawa-etal-2020-dro}. In this case, we apply group parity (up-sampling) on the training set to have group-balanced batches, while the optimization loss during training accounts for group-wise performance disparities using adaptive group-wise weights.
\setcounter{saveenum}{\value{enumi}}
\end{enumerate}

% \subsubsection*{Penalizing over-confidency}
\paragraph{Penalizing over-confidency} \emph{Over-confident} model predictions are considered an indication of bias based on the intuition that all simple feature correlations---leading to high confidence---are spurious \cite{gardner-etal-2021-competency}. We consider two methods from this line of work:
\vspace{-1mm}
\begin{enumerate}[align=left,leftmargin=*,wide = 0pt,itemsep=-2pt,label={\roman*})]
\setcounter{enumi}{\value{saveenum}}
    \item \textbf{Spectral Decoupling} (\textsc{fair-sd}) where the $L_2$ norm of the classification logits is used as a regularization penalty. The premise for this approach is that over-confidence reflects over-reliance to a limited number of relevant features, which leads to gradient starvation \cite{pezeshki2021gradient}. 
    \item \textbf{Debiased Focal Loss} (\textsc{fair-dfl})  where an additional task-agnostic classifier estimates if the model's prediction is going to be successful or not, and penalizes the model via focal loss~\cite{karimi-mahabadi-etal-2020-end} in case a successful outcome is highly predictable \cite{orgad-belinkov-2023-blind}.
\end{enumerate}

\noindent The first group of methods (representational bias) relies on demographic information, while the second group (penalizing over-confidency) is agnostic of demographic information, thus more easily applicable to different settings.

\subsection{Optimizing for Explainability}
\label{sec:opt-rationale}

We consider three alternative rationale extraction frameworks (REFs), where the models generate \emph{rationales}; \ie, a subset of the original tokens to predict the classification label. 
In these settings, the explanations (rationales) are \emph{faithful} by design, since the classifier (predictor) encodes only the rationales and has no access to the full text input, thus soley relies on those rationales at inference.  

\begin{enumerate}[align=left,leftmargin=*,wide = 0pt,itemsep=-2pt,label={\roman*})]
\item \textbf{Baseline} (\textsc{ref-base}) The baseline rationale extraction framework of \citet{lei-etal-2016-rationalizing} relies on two sub-networks (Eqs.~\ref{eq:1}-\ref{eq:4}): the \emph{rationale selector} that selects relevant input tokens to predict the correct label (Eq.~\ref{eq:1}-\ref{eq:2}), and the \emph{predictor} (Eq.~\ref{eq:3}-\ref{eq:4}) that predicts the classification task outcome based on the rationale provided by the first module.

\item \textbf{3-Player} (\textsc{ref-3p}) \citet{yu-etal-2019-rethinking} improved the aforementioned framework introducing a 3-player adversarial minimax game between the main predictor, the one using the rationale, and a newly introduced predictor using the complement of the rationale tokens. They found that this method improves classification performance, and the predicted rationales are more complete (\ie, they include a higher portion of the relevant information to solve the task) than the baseline framework.

\item \textbf{Rationale to Attention} (\textsc{ref-r2a}) More recently, \citet{yu2021understanding} introduced a new 3-player version where, during training, they minimize the performance disparity between the main predictor (the one using the rationales) and a second one using soft attention scores. They find this to result in rationales that better align with human rationales.

\end{enumerate}
% \vspace{-5mm}
\noindent For all examined rationale extraction frameworks, we use the latest implementations provided by \citet{yu2021understanding}, which use a top-$k$ token selector, instead of sparsity regularization~\cite{lei-etal-2016-rationalizing}:
\begin{eqnarray} 
    S &=& W^{H\times1} *\mathrm{TokenScorer}(X) \label{eq:1}\\
    Z &=& \mathrm{TopK}(X, S, k) \label{eq:2}\\
    R &=& Z * X \label{eq:3}\\
    L &=& \mathrm{Predictor}(R) \label{eq:4}
\label{eq:ref}
\end{eqnarray}

\noindent where $\mathrm{TokenScorer}$ and $\mathrm{Predictor}$ are Transformer-based language models (encoders), $X=[x_1, x_2, \cdots, x_n]$ are the input tokens, $S$ are the token importance scores based on the $\mathrm{TokenScorer}$ contextualized token representations, $Z$ is a binary mask representing which input tokens are the top-$k$ scored vs. the rest, $R$ is the rationale (masked version of the input tokens), and $L$ are the label logits. During training, the $\mathrm{TopK}$ operator is detached---since it is not differentiable---and gradients pass \emph{straight-through}~\cite{bengio-etal-2013} to the  $\mathrm{TokenScorer}$ to be updated.
For \textsc{ref-3p}, there is an additional adversarial $\mathrm{Predictor}$ (Eq.~\ref{eq:4}) which is fed with adversarial rationales ($R_{adv}$) based on the complement (\textsc{ref-3p}) of the original ones ($R$), while for \textsc{ref-r2a}, the adversarial predictor weighs the input tokens ($X$) given softmax-normalized scores ($S$).

\section{Experiments}

\subsection{Experimental Setup}

We fine-tune classifiers based on RoBERTa-base \cite{liu2019roberta} for all examined methods. In the case of the ECtHR dataset, which consists of long documents, we build hierarchical RoBERTa-based classifiers similar to \citet{chalkidis-etal-2022-lexglue}.\footnote{Similarly, rationales (Eq.~\ref{eq:1}-\ref{eq:3}) are computed based on paragraph-level, not token-level, representations.} We perform a grid search for the learning rate $\in[1e\!-\!5, 3e\!-\!5, 5e\!-\!5]$ with an initial warm-up of 10\%, followed by cosine decay, using AdamW~\citep{loshchilov2018decoupled}. We use a fixed batch size of 32 examples and fine-tune models up to 30 epochs with early stopping based on the validation performance. We fine-tune models with 5 different seeds and select the top-3 models (seeds) with the best overall validation performance (mF1) to report averaged results for all metrics. 

For methods optimizing for fairness, we rely on the LRP framework~\cite{ali-etal-2022-lrp} to extract post-hoc explanations, similar to~\citet{eberle-etal-2023-contrast-bios}.
\vspace{-2mm}
\paragraph{Evaluation metrics.}

Our main performance metric is macro-F1 (mF1); \ie, the F1 score macro-averaged across all classes, which better reflects the overall performance across classes regardless of their training support (robust to class imbalance). 

Regarding \emph{empirical fairness} metrics, we report group-wise performances (\eg, male and female mF1 in BIOS, and E.E. and the Rest in ECtHR) and their absolute difference (group disparity). Ideally, a fair(er) model will improve the worst-case performance, i.e., the lower performance across both groups, while reducing the group disparity.  

For \emph{explainability}, we report Area Over the Perturbation Curve (AOPC) for \emph{sufficiency} \cite{deyoung-etal-2020-eraser} as a proxy to \emph{faithfulness}~\cite{jacovi-goldberg-2020-towards}; \ie, how much explanations reflect the true reasoning---as reflected by importance scores---of a model. We compute sufficiency for all models using as a reference (classifier) a large RoBERTa model to have a fair common ground. We also report token-level recall at human level (R@k), similar to ~\citet{chalkidis-etal-2021-paragraph}, considering the top-$k$ tokens, where $k$ is the number of tokens annotated by humans,\footnote{In this case, all models are compared in a fair manner using the number of the selected tokens in the human rationale as an oracle.} as a metric of alignment (agreement) between model-based explanations and human rationales.

For estimating \emph{bias}, we report the $L_2$ norm of the classification logits, which is used as a regularization penalty by Spectral Decoupling~\cite{pezeshki2021gradient} as a proxy for confidence. We also report gender accuracy, as a proxy for bias, by fine-tuning probing classifiers on the protected attribute examined (\eg, gender classifiers for BIOS) initialized by the models previously fine-tuned on the downstream task (Section~\ref{sec:bias-fairness})\vspace{-1mm}

\subsection{Results on Synthetic Data}

In Table~\ref{tab:biased}, we present results for all fairness-promoting methods in the artificially unbalanced (biased) and balanced (debiased) versions of the BIOS dataset: \biased~and~\balanced~, described in Section~\ref{sec:datasets}. These can be seen as control experiments, to assess methods in edge cases. 
\vspace{-2mm}
\paragraph{Fairness methods rely on biases in data.} 
When training on \biased, we observe that all fairness-promoting methods outperform the baseline method in terms of our empirical fairness metrics: worst-group, i.e., female, performance and group disparity (difference in performance for male and female). We further see that almost all methods have mF1 scores of $0$ when it comes to \emph{male nurses} and very low scores ($15-49$) for \emph{female surgeons}. For both classes (\emph{nurse} and \emph{surgeon}), there were only their female and male counterpart, respectively, in the training dataset of \biased. This result suggests that all but one fairness-promoting methods (namely \textsc{fair-gn}) heavily rely on gender information to solve the task when such a spurious correlation is present. Only \textsc{fair-gn}, where gender information is neutralized is able to solve the task reliably, including almost no group disparity and mF1 scores $>60$ for male nurses and female surgeons. In Table~\ref{tab:words} in the Appendix, we present the top-attributed words for both occupations per gender which support this finding. 
All methods, except \textsc{fair-gn}, attribute gendered words a high (positive or negative) score following the imbalance in training.
Words such as `she', `mrs.', and `her' are positively attributed for females nurses, while `he' is negatively attributed for male nurses; and symmetrically the opposite for surgeons (Table~\ref{tab:words}). 
The only exception is \textsc{fair-gn}, in which case gender information has been neutralized during training and testing and the model can no longer exploit such superficial spurious correlations, \eg, that females can only be nurses or psychologists.
Concluding, all fairness-promoting methods \emph{improve} empirical fairness compared to the baseline, but in such extreme scenarios only a direct manual intervention on the data as in \textsc{fair-gn} reaches meaningful performance. 
\vspace{-2mm}
\paragraph{Data debiasing improves fairness methods.}
After downsampling the data to reach an equal number of males and females for all five professions for \balanced, we see almost equal performance across genders for \textsc{baseline}, \textsc{fair-gn} and \textsc{fair-dro} (\emph{lower} part of Table~\ref{tab:biased}).
Moreover, the performance for \textsc{fair-gn} and \textsc{fair-dro} is both higher and more equal across $M$ and $F$ than for \textsc{baseline}.
Overall, the models show an mF1 score of around $3\%$ lower than in the main results in Table~\ref{tab:results}, which is probably caused by down-sampling (fewer training samples), and to a smaller degree from not relying on gender bias.

\begin{table}[t]
    \centering
    \resizebox{\columnwidth}{!}{
        \begin{tabular}{lccc}
\toprule
     \multirow{2}{*}{\bf Method } & \multicolumn{3}{c}{Empirical Fairness (mF1)} \\
     & \bf \hspace{1MM}  M $\uparrow$ / F $\uparrow$ / Diff. $\downarrow$  & \bf Nurse (M) $\uparrow$ & \bf Surgeon (F) $\uparrow$\\
     \midrule
     \multicolumn{4}{c}{\biased~\emph{(Artificially Unbalanced)}} \\
     \midrule
     \textsc{baseline} & 45.9 / 34.6 / 11.3 & ~~0.0 & 14.8 \\
     \midrule
     \textsc{fair-gn}  &  \underline{81.7} / \underline{82.1} / ~~\underline{0.4} & \underline{61.5} & \underline{69.1} \\
     \textsc{fair-dro} & 53.5 / 60.6 / ~~7.1 & ~~0.0 & 48.5  \\
    \textsc{fair-sd}  & 48.7 / 50.5 / ~~1.8 & ~~0.0 & 38.7  \\
    \textsc{fair-dfl}  & 45.7 / 47.5 / ~~1.8 & ~~0.0 & 14.8  \\
    \midrule
    \multicolumn{4}{c}{\balanced~\emph{(Artificially Balanced)}} \\
    \midrule
     \textsc{baseline} & 83.6 / 84.4 / ~~0.8 & \underline{76.9} & 73.9 \\
     \midrule
     \textsc{fair-gn}  &  \underline{84.8} / 84.2 / ~~0.6 & 74.1 & 73.5\\
     \textsc{fair-dro} &  \underline{84.8} / \underline{85.0} / ~~\underline{0.2} & 74.1 & 79.2 \\
    \textsc{fair-sd}  &  83.5 / 86.2 / ~~2.6 & 71.4 & \underline{80.0} \\
    \textsc{fair-dfl}  &  82.6 / 85.8 / ~~3.2 & 74.1 & 76.6\\
    \bottomrule
\end{tabular}
    }
    \caption{Fairness-related metrics: macro-F1 (mF1) per group (male/female) and their absolute difference (Diff.), and worst-performing class (profession) per group, of fairness-promoting methods on the \emph{ultra-biased} or \emph{debiased} version of BIOS.}
    \label{tab:biased}
    \vspace{-5mm}
\end{table}

\begin{table*}[t]
    \centering
    \resizebox{\textwidth}{!}{
        \begin{tabular}{lcccc|c|cccc}
\toprule
    \bf {}  & \multicolumn{4}{c|}{\bf BIOS -- Occupation Classification} &&  \multicolumn{4}{c}{\bf ECtHR -- ECHR Violation Prediction} \\
     \cmidrule{2-5}
     \cmidrule{7-10}
     \multirow{2}{*}{\bf Method }  & \multirow{2}{*}{\bf mF1} & Empirical Fairness & \multicolumn{2}{c|}{Explainability} &&  \multirow{2}{*}{\bf mF1} & Empirical Fairness & \multicolumn{2}{c}{Explainability} \\
     && \bf mF1 (M / F / Diff.) & \bf AOPC &  \bf R@k & &&  \bf mF1 (EE / R / Diff.) & \bf AOPC &  \bf R@k\\
     \midrule
     \textsc{baseline} & \underline{\textbf{88.1}} & 85.5 / \textbf{87.5} / 2.0 & \underline{88.5} & \textbf{52.0}  & & 83.5 & 83.1 / 83.3 / \textbf{0.2} & 77.4 & 48.8 \\
     \midrule
     \multicolumn{10}{c}{\emph{Optimizing for Fairness}} \\
     \midrule
     \textsc{fair-gp}  & 87.8 & 83.8 / \underline{\textbf{87.5}} / 3.7 & 88.0 & 47.8 & & 83.9  & 83.5 / 81.8 / 2.5 & 77.0 & \underline{50.5} \\
     \textsc{fair-gn}  & 87.8 & 82.5 / 86.8 / 4.2 & 88.0 & 48.7  & & \multicolumn{4}{c}{------------ Not Applicable (N/A)\footnotemark[4] ------------}\\
     \textsc{fair-dro} & 87.6 & 84.2 / 86.4 / 2.2 & 88.4 & 48.8 & &  83.9  & 83.6 / 80.6 / 3.0 & 77.9 & 49.8  \\
    \textsc{fair-sd}  & \underline{87.9} & \underline{85.6} / 86.6 / \underline{\textbf{1.0}} & \underline{88.5} & \underline{49.4} & &  \underline{\textbf{84.9}} & \underline{\textbf{84.2}} / \underline{\textbf{87.1}} / 2.9 & \underline{\textbf{78.8}} & 49.9  \\
     \textsc{fair-dfl}  & 87.6 & 84.5 / 86.4 / 1.9 & 87.3 & 45.5 & & 84.3 & 84.1 / 83.6 / \underline{0.5} & 78.2 & 49.2  \\
     \midrule
     \multicolumn{10}{c}{\emph{Optimizing for Explainability}} \\
     \midrule
     \textsc{ref-base} & 85.3 & 82.2 / 83.9 / \underline{1.7} & 78.1 & 45.7 & & 81.8  & 81.9  / 81.3  / \underline{0.6}  & 73.2 & \underline{\textbf{57.1}}\\ 
     \textsc{ref-3p}   & \underline{86.4} & 81.8 / 85.0 / 3.1 & 79.6 & 44.3  & & 83.1  & \underline{83.3}  / 80.8  / 2.5  & 73.3  & 54.0\\
     % \textsc{ref-rand}  & 86.3 & 82.4 / 85.4 / 3.0 & 80.4 & 44.9  & & 00.0  & 00.0  / 00.0  / 0.0  & 00.0  & 00.0\\
     \textsc{ref-r2a}  & 86.1 & \underline{82.4} / \underline{85.4} / 3.0 & \underline{82.9} & \underline{50.7} & & 82.8  & 82.6  / \underline{83.4}  / 0.8  & \underline{74.5}  & 50.9\\
     % \midrule
     % \multicolumn{10}{c}{\emph{Optimizing for Both}} \\
     % \midrule
     % \textsc{ref-base-sd}  &  \multicolumn{4}{c|}{------------ Results not in time ------------} & & 82.0 & 81.9 / 81.3 / 0.6 & 72.6  & 53.8\\
     % \textsc{ref-r2a-sd}  &  \multicolumn{4}{c|}{------------ Results not in time ------------} & & \underline{82.4} & \underline{82.1} / \underline{82.1} / \textbf{0.0} & 74.3  & 52.9 \\
    \bottomrule
\end{tabular}
    }
    \caption{Test Results for all examined methods. We report the overall macro-F1 (mF1), alongside fairness-related metrics: macro-F1 (mF1) per group and their absolute difference (Diff.), also referred to as group disparity; and explainability-related scores: AOPC for faithfulness and token R@k for human-model rationales alignment. The best scores across all models in the same group (\textsc{fair-}, \textsc{ref-}) are \underline{underlined}, and the best scores overall are in \textbf{bold}. We present detailed validation and test results including standard deviations in Tables~\ref{tab:val_results}-~\ref{tab:xai_all}.}
    \label{tab:results}
    \vspace{-3mm}
\end{table*}

\begin{figure*}[t]
    \centering
    \resizebox{0.95\textwidth}{!}{
    \includegraphics{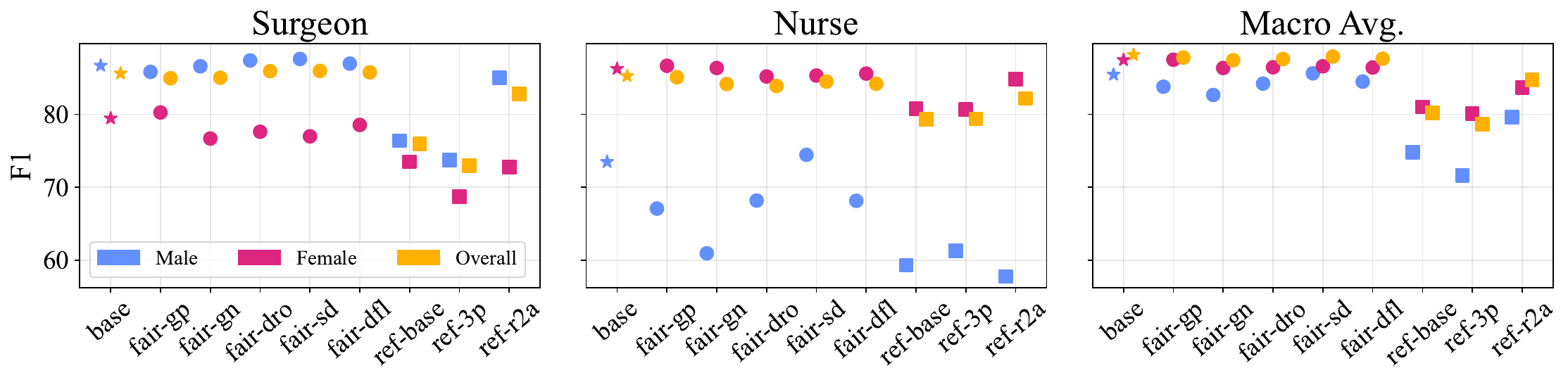}
    }
    \vspace{-3mm}
    \caption{F1 and macro-F1 scores for the classes \emph{surgeon} and \emph{nurse} from the BIOS dataset for all methods per gender. Baseline is marked as $\star$, fairness-promoting methods as $\circ$, and REFs as $\square$.  We see a severe drop in performance for the underrepresented class (female surgeons and male nurses). }
    \label{fig:bios-plot-mini}
    \vspace{-3mm}
\end{figure*}

\subsection{Main Results on Real Data}
\label{sec:main_results}

In Table~\ref{tab:results}, we present results for all examined methods for both datasets, BIOS and ECtHR. 
\vspace{-2mm}
\paragraph{Overall performance.} In the case of BIOS, we observe a drop in performance, in particular when optimizing for explainability where mF1 scores decrease from $88\%$ down to $85\%$ in comparison to the \textsc{baseline}. We also see an increase in group disparity for 3 out of 5 fairness-promoting methods and 2 out of 3 explainability-promoting methods. This is further supported by the results in Figure~\ref{fig:bios-plot-mini}, where we show F1 scores for the two classes \emph{surgeon} and \emph{nurse} from the BIOS dataset (see Figure~\ref{fig:bios-plot} in Appendix for results across all classes and metrics). We see a severe drop in performance for the two most underrepresented demographics, female surgeons and male nurses, of up to $25\%$ in comparison to the overrepresented counterpart.
In contrast, in the case of ECtHR, fairness-promoting (bias mitigation) methods, have a beneficial effect, especially in the case of confidence-related methods \textsc{fair-sd} and \textsc{fair-dfl} where overall task performance increase by $0.8-1.4\%$ with respect to the \textsc{baseline}. We suspect that the positive impact in the case of ECtHR is partly a side-effect of a higher class imbalance (label-wise disparity), e.g., there are many more cases tagged with Art.~6 compared to the rest of the labels, as demonstrated in Table~\ref{tab:bios_stats} (lower part), similar to the findings of~\citet{chalkidis-sogaard-2022-improved} who showed that \textsc{fair-sd} works particularly well for high class imbalance.

\paragraph{Fairness-promoting methods.}  
In the case of BIOS, we observe that only \textsc{fair-sd} can slightly improve empirical fairness, reflected through lower group disparity at the cost of a lower group performance for \textsc{female (F)}, while the remaining fairness-promoting methods lead to a more or similar unfair performance. We observe similar results for ECtHR, where only two out of four methods (\textsc{fair-sd}, \textsc{fair-dfl}) are able to improve the performance for both groups (EE, R), while increasing the group disparity, as all other methods.\footnote{We do not consider \textsc{fair-gn} in ECtHR, since there is no straightforward way to anonymize (neutralize) information relevant to the defendant state, which is potentially presented in the form of mentions to locations, organization, etc..}  Concluding, we find that bias mitigation algorithms do not always lead to fairer models which is in line with \citet{cabello2023independence}. Considering explainability-related metrics---faithfulness and human-model alignment as measured by R@k---for the fairness-promoting (bias mitigation) methods,
% \footnote{We remind readers that in this case, LRP is used to estimate relevance (importance) attribution scores.} 
we observe that improved empirical fairness does not lead to \emph{better} model explanations, neither for faithfulness (AOPC) nor for plausibility (R@k) when comparing \textsc{fair-sd} and \textsc{fair-dfl} with the \textsc{baseline}.

\paragraph{Rationale Extraction Frameworks (REFs).} 
Considering the results for the rationale extraction frameworks (REFs, see Section~\ref{sec:opt-rationale}) presented in the lower part of Table~\ref{tab:results}, we observe that the overall performance (mF1) decreases by~2-3\% in the case of BIOS, and by~0.5-2\% for ECtHR, since the models' predictor only considers a subset of the original input, the rationale. All REFs that aim to improve explainability compromise empirical fairness (\ie, performance disparity) in both datasets. 

When it comes to explainability, the results are less clear. For BIOS, both scores---faithfulness and human-model alignment---, drop in comparison to the baseline, while all REF methods substantially improve human--model alignment (by 2-8\%) in the case of ECtHR. 
For REFs, we also observe that an improvement in empirical fairness does not correlate with an improvement in explainability. 
\vspace{-2mm}
\subsection{Bias Mitigation $\neq$ Empirical Fairness}
\label{sec:bias-fairness}

Based on our findings in Section~\ref{sec:main_results}, we investigate the dynamics between bias mitigation and empirical fairness further. We examine the fairness-promoting methods on both datasets considering two indicators of bias: (a) the $L2$ norm of the classification logits as a proxy for the model's over-confidency (also used as a penalty term by \textsc{fair-sd}), and (b) the accuracy of a probing classifier for predicting the attribute (gender/nationality). This probing classifier relies on a frozen encoder that was previously fine-tuned on the occupation/article classification task with a newly trained classification head. To avoid conflating bias with features learned for the main classification tasks, e.g., medical occupation classification for BIOS, we use new datasets, excluding documents with the original labeling, e.g., for BIOS we use 3K biographies for 5 non-medical professions to train the gender classifier. With this analysis, we want to quantify to what degree we can extract information on gender/nationality, from the biographies/court cases.

\begin{table}[t]
    \centering
    \resizebox{0.9\columnwidth}{!}{
        \begin{tabular}{lcccc}
\toprule
     \multirow{2}{*}{\bf Method}  & \multicolumn{2}{c}{Fairness (mF1)} & \multicolumn{2}{c}{Bias Proxies} \\
     & \bf WC $\uparrow$ & \bf Diff.  $\downarrow$ & \bf $|L2|$    $\downarrow$& \bf Group Acc.  $\downarrow$ \\
     \midrule
     \multicolumn{5}{c}{\bf BIOS – Occupation Classification} \\
     \midrule
     \textsc{baseline} & 85.5 & 2.0 & 12.6 & 93.2 \\
     \midrule
     \textsc{fair-gp}  & 83.8 & 3.7 & 18.6 & 96.6 \\
     \textsc{fair-gn}  & 82.5 & 4.2 & 11.6 & \underline{65.4} \\
     \textsc{fair-dro} & 84.2 & 2.2 & 21.2 & 98.2 \\
    \textsc{fair-sd}  & \underline{85.6} & \underline{1.0} & \underline{00.7} & 96.0\\
     \textsc{fair-dfl}  & 84.5 & 1.9 & 06.5 & 96.2 \\
     % \midrule
     %  \textsc{ref-base} & 82.2 & \underline{1.7} & 05.3 & 66.2 \\ 
     % \textsc{ref-3p}   & 81.8 & 3.1 & 05.2 & 65.8 \\
     % % \textsc{ref-rand}  & \underline{82.4} & 3.0 & 05.4 & 59.7 \\
     % \textsc{ref-r2a}  & \underline{82.4} & 3.0 & \underline{00.3} & \underline{55.0} \\
     \midrule
     \multicolumn{5}{c}{\bf ECtHR – ECHR Violation Prediction} \\
     \midrule
     \textsc{baseline} & 83.1 & \underline{0.2} & 10.7 & 75.0 \\
     \midrule
     \textsc{fair-gp} & 81.8  & 2.7 & 11.3 & 69.6 \\
     \textsc{fair-dro} & 80.6 & 3.0 & 16.7 & 76.2 \\
     \textsc{fair-sd} & \underline{84.2} & 2.9 & \underline{00.4} & 72.4 \\
     \textsc{fair-dfl} & 83.6 & 0.5 & 04.5 & \underline{63.0} \\
     % \midrule
     % \textsc{ref-base} & 81.3 & 0.6 & 13.1 & - \\
     % \textsc{ref-3p} & 80.8 & 2.5 & 13.5 & - \\
     % \textsc{ref-r2a} & 82.6 & 0.8 & 12.2  & - \\
     
    \bottomrule
\end{tabular}
    }
    \caption{Fairness- and bias-related metrics. We show again downstream task performance for \emph{Worst-Case} (WC) and the group-wise difference as indicators for empirical fairness. We further add $L2$ norm of the classification logits as an indicator for (over-)confidency and accuracy for group classification both as bias proxies.}
    \label{tab:bias_indicators}
    \vspace{-5mm}
\end{table}

In Table~\ref{tab:bias_indicators}, we report empirical fairness metrics and the bias indicators (proxies) for all examined methods together with F1 scores for \emph{worst-case-scenario} (WC) across all classes and the difference in mF1 between the two groups from Table~\ref{tab:results}. First of all, with respect to BIOS, we observe that all fairness-promoting algorithms, except \textsc{fair-gn}, show a high accuracy for gender classification ($>95\%$), thus, are more biased compared to the baseline with respect to gender classification accuracy. Furthermore, the least biased classifier (\textsc{fair-gn}), is outperformed by all other fairness-promoting methods in both empirical fairness metrics.
In the case of ECtHR, we observe that 3 out of 4 fairness-promoting methods decrease bias, shown by lower group accuracy and lower confidency scores (L2 norm) for \textsc{fair-sd} and \textsc{fair-dfl}. This does not lead to improvements in empirical fairness, with the exception of worst-case performance for \textsc{fair-sd} and \textsc{fair-dfl}.
\vspace{-2mm}
\section{Conclusion}
\vspace{-1mm}
We systematically investigated the interplay between empirical fairness and explainability, two key desired properties required for trustworthy NLP systems.
We did so by considering five fairness-promoting methods, and three rationale extraction frameworks, across two datasets for multi-class classification (BIOS and ECtHR).
Based on our results, we see that improving either empirical fairness or explainability does \emph{not} improve the other. Interestingly, many fairness-promoting methods do not mitigate bias, nor promote fairness as intended, while we find that these two dimensions are also orthogonal (Figure~\ref{fig:enter-label}).
% , similar to the recent findings of \citet{cabello2023independence}. 
Furthermore, we see that (i) gender information is encoded to a high amount in the occupation classification task, and (ii) the only successful strategy to prevent this, seems to be the normalization across gender during training. We conclude that trustworthiness, reflected through empirical fairness and explainability, is still an open challenge.
With this work, we hope to encourage more efforts that target a holistic investigation and the development of new algorithms that promote both crucial qualities.
% at the same time.

\section*{Limitations}
Our analysis is limited to English text classification datasets. In order to make general conclusions about the interplay between fairness and explainability, one need to extend this analysis to other languages and downstream tasks. 

Furthermore, we argue within the limited scope of specific definitions of fairness, bias and explainability for binary attributes. This analysis could be applied to further attributes. Also, we have not included human evaluation with respect to explainability, i.e., humans evaluating the rationales for usability and plausibility, see \citet{brandl2022evaluating, yin-neubig-2022-interpreting}.

\section*{Acknowledgements}
We thank our colleagues at the CoAStaL NLP group for fruitful discussions in the beginning of the project. In particular, we would like to thank Daniel Hershcovich, Desmond Elliott and Laura Cabello for valuable comments on the manuscript. 
{\scriptsize\euflag} EB is supported by the funding from the European Union's Horizon 2020 research and innovation programme under the Marie Sk\l{}odowska-Curie grant agreement No 801199.
IC is funded by the Novo Nordisk Foundation (grant NNF 20SA0066568). SB receives funding from the European Union under the Grant Agreement no.~10106555, FairER. Views and opinions expressed are those of the author(s) only and do not necessarily reflect those of the European Union or European Research Executive Agency (REA). Neither the European Union nor REA can be held responsible for them.

\bibliography{anthology,custom}
\bibliographystyle{acl_natbib}

\appendix

\begin{table*}[t]
    \centering
    \resizebox{\textwidth}{!}{
        \begin{tabular}{lcccc|c|cccc}
\toprule
    \bf {}  & \multicolumn{4}{c|}{\bf BIOS -- Occupation Classification} &&  \multicolumn{4}{c}{\bf ECtHR -- ECHR Violation Prediction} \\
     \cmidrule{2-5}
     \cmidrule{6-10}
     \multirow{2}{*}{\bf Method }  & \multirow{2}{*}{\bf Avg.} & \multicolumn{3}{c|}{Empirical Fairness} &&  \multirow{2}{*}{\bf Avg.} & \multicolumn{3}{c}{Empirical Fairness} \\
     && \bf M & \bf F & \bf Diff. & &&  \bf EE & \bf R & \bf Diff. \\
     \midrule
\textsc{baseline} & 89.7 $\pm$ \small{0.1} & \underline{\textbf{90.9}}  $\pm$ \small{0.2} & 86.2  $\pm$ \small{0.7} & 4.7  $\pm$ \small{0.7} && 87.2 $\pm$ \small{0.2} & 87.4  $\pm$ \small{0.7} & 84.3  $\pm$ \small{3.4} & 3.1  $\pm$ \small{1.7} \\
     \midrule
     \multicolumn{6}{c}{\emph{Optimizing for Fairness}} \\
     \midrule
\textsc{fair-gp} & 89.9 $\pm$ \small{0.1} & 89.7  $\pm$ \small{1.0} & 86.9  $\pm$ \small{0.1} & 2.8  $\pm$ \small{1.1} && 86.3 $\pm$ \small{0.4} & 87.0  $\pm$ \small{0.4} & 81.4  $\pm$ \small{0.8} & 5.6  $\pm$ \small{0.5} \\
\textsc{fair-gn} & 89.1 $\pm$ \small{0.2} & 86.7  $\pm$ \small{1.3} & 85.7  $\pm$ \small{1.0} & 1.0  $\pm$ \small{1.4} && \multicolumn{4}{c}{------------ Not Applicable (N/A) ------------} \\
\textsc{fair-dro} & 89.7 $\pm$ \small{0.3} & 90.5  $\pm$ \small{1.0} & 86.4  $\pm$ \small{0.8} & 4.1  $\pm$ \small{1.7} && 86.9 $\pm$ \small{0.9} & 87.6  $\pm$ \small{0.7} & 82.5  $\pm$ \small{2.4} & 5.1  $\pm$ \small{1.8} \\
\textsc{fair-sd} & \underline{\textbf{90.3}} $\pm$ \small{0.0} & 90.2  $\pm$ \small{0.9} & 87.7  $\pm$ \small{0.3} & 2.5  $\pm$ \small{0.6} && 87.6 $\pm$ \small{1.1} & \underline{\textbf{88.5}}  $\pm$ \small{1.0} & 82.9  $\pm$ \small{1.9} & 5.6  $\pm$ \small{1.0} \\
\textsc{fair-dfl} & 90.0 $\pm$ \small{0.1} & 88.5  $\pm$ \small{0.6} & \underline{\textbf{88.0}}  $\pm$ \small{0.4} & \underline{\textbf{0.5}}  $\pm$ \small{0.7} && \underline{\textbf{88.1}} $\pm$ \small{0.7} & 88.4  $\pm$ \small{0.8} & \underline{\textbf{85.8}}  $\pm$ \small{2.9} & \underline{\textbf{2.6}}  $\pm$ \small{2.9} \\
     \midrule
     \multicolumn{10}{c}{\emph{Optimizing for Explainability}} \\
     \midrule
     \textsc{ref-base} & 87.2 $\pm$ \small{0.2} & \underline{88.5}  $\pm$ \small{0.2} & 82.7  $\pm$ \small{1.2} & 5.8  $\pm$ \small{1.1} && 87.1 $\pm$ \small{0.2} & 87.5  $\pm$ \small{0.2} & 85.1  $\pm$ \small{2.5} & 3.1  $\pm$ \small{1.8} \\
\textsc{ref-3p} & 86.8 $\pm$ \small{0.6} & 87.1  $\pm$ \small{2.1} & 81.1  $\pm$ \small{0.9} & 6.0  $\pm$ \small{1.4} && 86.9 $\pm$ \small{0.5} & 87.7  $\pm$ \small{0.3} & 83.7  $\pm$ \small{1.9} & 4.1  $\pm$ \small{2.0} \\
\textsc{ref-r2a} & \underline{87.5} $\pm$ \small{0.4} & \underline{88.5}  $\pm$ \small{1.5} & \underline{83.7}  $\pm$ \small{1.3} & \underline{4.8}  $\pm$ \small{1.9} && \underline{88.0} $\pm$ \small{0.9} & \underline{88.4}  $\pm$ \small{0.8} & \underline{85.8}  $\pm$ \small{0.9} & \underline{2.6}  $\pm$ \small{0.3} \\
    \bottomrule
\end{tabular}
    }
    \caption{Validation Results (mF1) with standard deviations ($\pm$) for all examined methods in the examined datasets. We report the overall (Avg.) macro-F1 (mF1), alongside fairness-related metrics: macro-F1 (mF1) per group and their absolute difference (Diff.), also referred to as group disparity.  The best scores across all models in the same group (\textsc{fair-}, \textsc{ref-}) are \underline{underlined}, and the best scores overall are in \textbf{bold}.}
    \label{tab:val_results}
    \vspace{-3mm}
\end{table*}

\begin{table*}[t]
    \centering
    \resizebox{\textwidth}{!}{
        \begin{tabular}{lcccc|c|cccc}
\toprule
    \bf {}  & \multicolumn{4}{c|}{\bf BIOS -- Occupation Classification} &&  \multicolumn{4}{c}{\bf ECtHR -- ECHR Violation Prediction} \\
     \cmidrule{2-5}
     \cmidrule{6-10}
     \multirow{2}{*}{\bf Method }  & \multirow{2}{*}{\bf Avg.} & \multicolumn{3}{c|}{Empirical Fairness} &&  \multirow{2}{*}{\bf Avg.} & \multicolumn{3}{c}{Empirical Fairness} \\
     && \bf M & \bf F & \bf Diff. & &&  \bf EE & \bf R & \bf Diff. \\
     \midrule
\textsc{baseline} & \textbf{88.1} $\pm$ 0.3 & 85.5  $\pm$ 1.4 & \textbf{87.5}  $\pm$ 0.9 & 2.0  $\pm$ 1.2 && 83.5 $\pm$ 0.6 & 83.1  $\pm$ 0.7 & 83.3  $\pm$ 0.8 & 0.2  $\pm$ 0.7 \\
     \midrule
     \multicolumn{6}{c}{\emph{Optimizing for Fairness}} \\
     \midrule
\textsc{fair-gp} & 87.8 $\pm$ 0.4 & 83.8  $\pm$ 1.6 & \underline{\textbf{87.5}}  $\pm$ 0.3 & 3.7  $\pm$ 1.2 && 83.9 $\pm$ 0.2 & 83.5  $\pm$ 0.2 & 81.8  $\pm$ 2.2 & 2.5  $\pm$ 1.3 \\
\textsc{fair-gn} & 87.8 $\pm$ 0.2 & 82.5  $\pm$ 0.6 & 86.8  $\pm$ 0.6 & 4.2  $\pm$ 1.1 && \multicolumn{4}{c}{------------ Not Applicable (N/A) ------------} \\
\textsc{fair-dro} & 87.6 $\pm$ 0.6 & 84.2  $\pm$ 0.4 & 86.4  $\pm$ 1.2 & 2.2  $\pm$ 1.3 && 83.9 $\pm$ 0.5 & 83.6  $\pm$ 0.5 & 80.6  $\pm$ 2.0 & 3.0  $\pm$ 1.7 \\
\textsc{fair-sd} & \underline{87.9} $\pm$ 0.1 & \underline{\textbf{85.6}}  $\pm$ 0.3 & 86.6  $\pm$ 0.2 & \underline{1.0}  $\pm$ 0.4 && \underline{\textbf{84.9}} $\pm$ 0.2 & \underline{\textbf{84.2}}  $\pm$ 0.2 & \underline{\textbf{87.1}}  $\pm$ 2.9 & 2.9  $\pm$ 3.1 \\
\textsc{fair-dfl} & 87.6 $\pm$ 0.6 & 84.5  $\pm$ 0.8 & 86.4  $\pm$ 0.6 & 1.9  $\pm$ 0.9 && 84.3 $\pm$ 1.0 & 84.1  $\pm$ 0.6 & 83.6  $\pm$ 4.2 & 0.5  $\pm$ 1.8 \\
     \midrule
     \multicolumn{10}{c}{\emph{Optimizing for Explainability}} \\
     \midrule
     \textsc{ref-base} & 85.3 $\pm$ 0.9 & 82.2  $\pm$ 1.1 & 83.9  $\pm$ 0.9 & \underline{1.7}  $\pm$ 1.0 && 81.8 $\pm$ 1.8 & 81.9  $\pm$ 2.1 & 81.3  $\pm$ 3.5 & \underline{0.6}  $\pm$ 0.9 \\
\textsc{ref-3p} & \underline{86.4} $\pm$ 0.7 & 81.8  $\pm$ 1.0 & 85.0  $\pm$ 1.4 & 3.1  $\pm$ 1.4 && \underline{83.1} $\pm$ 0.3 & \underline{83.3}  $\pm$ 0.6 & 80.8  $\pm$ 2.2 & 2.5  $\pm$ 1.8 \\
\textsc{ref-r2a} & 86.1 $\pm$ 0.6 & \underline{82.4}  $\pm$ 0.4 & \underline{85.4}  $\pm$ 1.0 & 3.0  $\pm$ 1.0 && 82.8 $\pm$ 0.6 & 82.6  $\pm$ 0.5 & \underline{83.4}  $\pm$ 2.6 & 0.8  $\pm$ 0.8 \\
    \bottomrule
\end{tabular}
    }
    \caption{Test Results (mF1)  with standard deviations ($\pm$) for all examined methods in the examined datasets. We report the overall (Avg.) macro-F1 (mF1), alongside fairness-related metrics: macro-F1 (mF1) per group and their absolute difference (Diff.), also referred to as group disparity.  The best scores across all models in the same group (\textsc{fair-}, \textsc{ref-}) are \underline{underlined}, and the best scores overall are in \textbf{bold}.}
    \label{tab:test_results}
    \vspace{-3mm}
\end{table*}

\begin{table*}[t]
    \centering
    \resizebox{0.8\textwidth}{!}{
        \begin{tabular}{lcc|c|cc}
\toprule
    \bf {}  & \multicolumn{2}{c|}{\bf BIOS -- Occupation Classification} &&  \multicolumn{2}{c}{\bf ECtHR -- ECHR Violation Prediction} \\
     \cmidrule{2-3}
     \cmidrule{4-6}
     \multirow{2}{*}{\bf Method } & \multicolumn{2}{c|}{Explainability} && \multicolumn{2}{c}{Explainability} \\
     & \hspace{7mm} \bf AOPC \hspace{7mm} &  \hspace{7mm} \bf R@k \hspace{7mm} && \hspace{7mm} \bf AOPC \hspace{7mm} &  \hspace{7mm} \bf R@k \hspace{7mm} \\
     \midrule
     \textsc{baseline} & \textbf{88.5} $\pm$ 0.0 & \textbf{52.0} $\pm$ 1.7 && 77.4 $\pm$ 0.8 & 48.8 $\pm$ 0.2\\
     \midrule
     \multicolumn{6}{c}{\emph{Optimizing for Fairness}} \\
     \midrule
     \textsc{fair-gp} & 88.0 $\pm$ 0.0 & 47.8 $\pm$ 2.5 && 77.0 $\pm$ 0.7 & 50.5 $\pm$ 0.4\\
     \textsc{fair-gn} & 88.0 $\pm$ 0.0 & 48.7 $\pm$ 2.3 && \multicolumn{2}{c}{------ Not Applicable (N/A) -------}\\
     \textsc{fair-dro} & 88.4 $\pm$ 0.0 & 48.8 $\pm$ 0.9 && 77.9 $\pm$ 0.2 & 49.8 $\pm$ 0.8\\
     \textsc{fair-sd} & \underline{\textbf{88.5}} $\pm$ 0.0 & \underline{49.4} $\pm$ 3.2 && \underline{\textbf{78.8}} $\pm$ 0.8 & 49.9 $\pm$ 0.3\\
     \textsc{fair-dfl} & 87.3 $\pm$ 0.0 & 45.5 $\pm$ 2.4 && 78.2 $\pm$ 0.7 & 49.2 $\pm$ 1.6\\
     \midrule
     \multicolumn{6}{c}{\emph{Optimizing for Explainability}} \\
     \midrule
     \textsc{ref-base} & 78.1 $\pm$ 0.0 & 45.7 $\pm$ 4.0 && 73.2  $\pm$ 1.4 & \underline{\textbf{57.1}} $\pm$ 0.7\\
     \textsc{ref-3p} & 79.6 $\pm$ 0.0 & 44.3 $\pm$ 2.9 && 73.3 $\pm$ 0.5 & 54.0 $\pm$ 1.0\\
     \textsc{fair-r2a} & \underline{82.9} $\pm$ 0.0 & \underline{50.7} $\pm$ 7.4 && \underline{74.9} $\pm$ 1.0 & 50.9 $\pm$ 0.3\\
    \bottomrule
\end{tabular}
    }
    \caption{Test Results for all examined methods. We report explainability-related scores  with standard deviations ($\pm$): AOPC for faithfulness and token R@k for human-model rationales alignment. The best scores across all models in the same group (\textsc{fair-}, \textsc{ref-}) are \underline{underlined}, and the best scores overall are in \textbf{bold}.}
    \label{tab:xai_all}
    \vspace{-3mm}
\end{table*}

\section{More results}
\label{sec:appendix}

In Table~\ref{tab:val_results}-~\ref{tab:test_results}, we present validation, and test results across all methods for both examined datasets.

\begin{figure*}[t]
    \centering
    \resizebox{\textwidth}{!}{
    \includegraphics{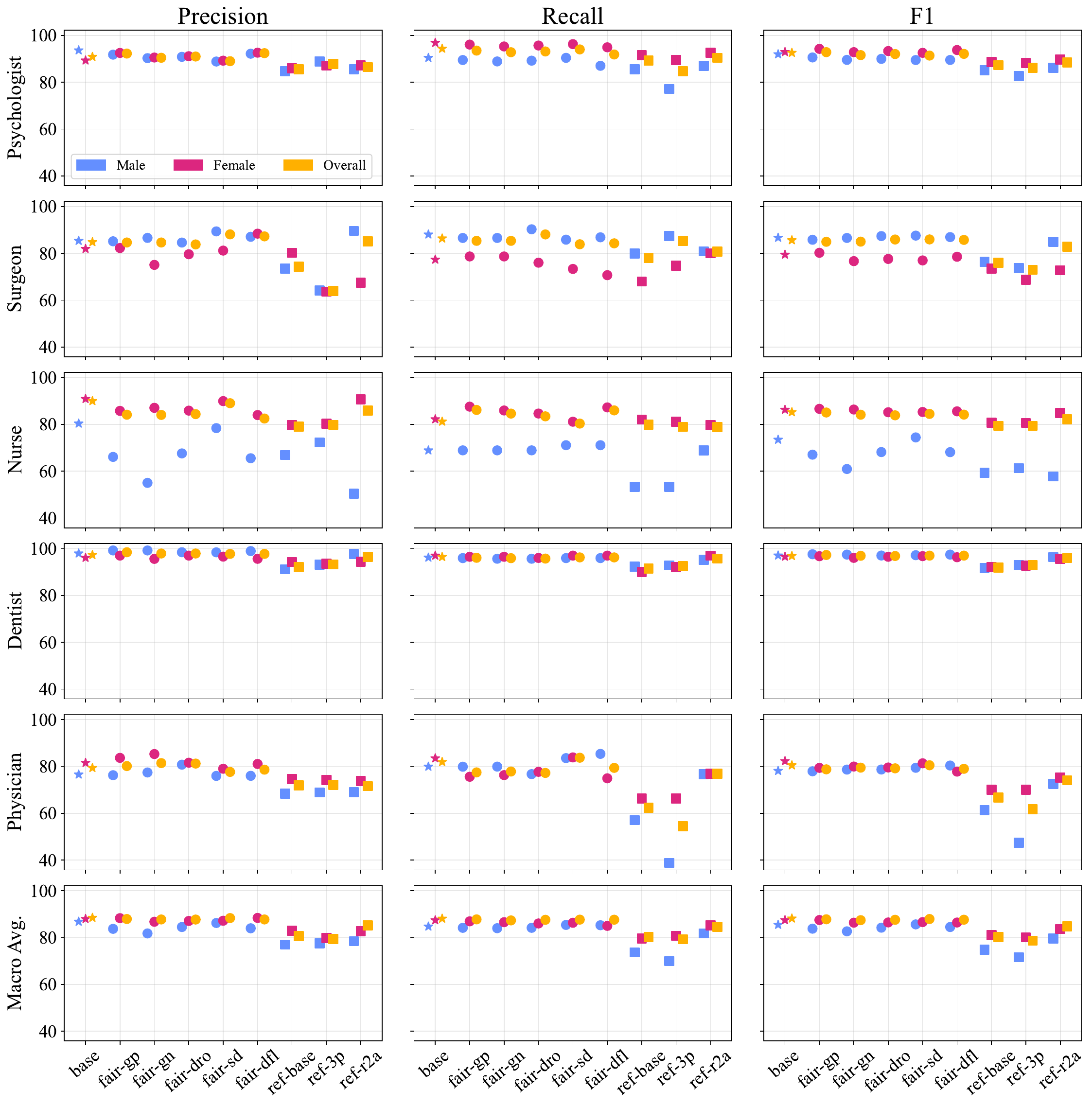}
    }
    \vspace{-3mm}
    \caption{Precision, Recall, and F1 across different medical occupations of the BIOS dataset for both (male, female) genders. A smaller gap between male (blue) and female (orange) performance represents a ``fairer'' model.}
    \vspace{-3mm}
    \label{fig:bios-plot}
\end{figure*}

In Table \ref{tab:words}, we present the list of words that were assigned the highest importance scores (positive and negative) for the 5 fairness-promoting methods and the baseline on the BIOS dataset. Additionally, we show class-wise F1 scores, separated by gender, for the BIOS dataset in Figure \ref{fig:bios-plot}.

\begin{table*}[t]
    \centering
    \resizebox{\textwidth}{!}{
    \begin{tabular}{l|c|c|c|c|c|c|c|c}
        \multirow{3}{*}{\bf Method} & \multicolumn{4}{c|}{\textsc{Nurse}}  & \multicolumn{4}{c}{ \textsc{Surgeon}} \\ 
         & \multicolumn{2}{c|}{\textsc{Positive}} & \multicolumn{2}{c|}{\textsc{Negative}} & \multicolumn{2}{c|}{\textsc{Positive}} & \multicolumn{2}{c}{\textsc{Negative}}    \\
         & M & F & M & F & M & F & M & F \\
        \midrule
         \multirow{5}{*}{BASELINE} & (nursing, 0.2) & (\textbf{mrs.}, 0.4) & (\textbf{he}, -0.3) & (research, -0.2) & (surgeon, 0.4) & (surgery, 0.5) & (working, -0.2) & (\textbf{she}, -0.3) \\ 
& (nurse, 0.2) & (nurses, 0.4) & - & (inc, -0.2) & (surgery, 0.4) & (practice, 0.1) & (care, -0.2) & (\textbf{her}, -0.1) \\ 
& - & (nursing, 0.3) & - & (no, -0.1) & (surgical, 0.3) & (dr., 0.1) & (interests, -0.2) & (health, -0.1) \\ 
& - & (\textbf{she}, 0.3) & - & (\textbf{elizabeth}, -0.1) &(surgeons, 0.3) & (treatment, 0.1) & (health, -0.1) & - \\ 
& - & (nurse, 0.2) & - & (mental, -0.1) & (neurosurgery, 0.2) & - & (md, -0.1) & - \\ 
\midrule
\multirow{5}{*}{FAIR-GN} & (nurse, 0.5) & (nurse, 0.6) & - & (research, -0.2) &(surgeon, 0.5) & (surgery, 0.6) & (working, -0.2) & (health, -0.2) \\ 
& (nursing, 0.4) & (nursing, 0.4) & - & (dr., -0.1) & (neurosurgery, 0.4) & (dr., 0.1) & (group, -0.2) & (center, -0.1) \\ 
& - & (nurses, 0.4) & - & (practice, -0.1) &(surgery, 0.4) & - & (over, -0.1) & - \\ 
& - & (rn, 0.3) & - & (work, -0.1) & (surgeons, 0.4) & - & (health, -0.1) & - \\ 
& - & (diabetes, 0.1) & - & - &(surgical, 0.3) & - & (general, -0.1) & - \\  
\midrule
\multirow{5}{*}{FAIR-DRO} & (nurse, 0.1) & (\textbf{mrs.}, 0.4) & (\textbf{he}, -0.2) & (research, -0.3) &(surgeon, 0.4) & (surgery, 0.5) & (care, -0.2) & (\textbf{she}, -0.3) \\ 
& (nursing, 0.1) & (nursing, 0.3) & - & (mental, -0.2) &(surgeons, 0.4) & - & (group, -0.2) & (\textbf{her}, -0.2) \\ 
& - & (\textbf{she}, 0.3) & - & (affiliates, -0.1) &(surgery, 0.3) & - & (5, -0.2) & (health, -0.2) \\ 
& - & (nurses, 0.2) & - & (no, -0.1) &(neurosurgery, 0.3) & - & (areas, -0.1) & - \\ 
& - & (\textbf{ms.}, 0.2) & - & (without, -0.1) &(surgical, 0.3) & - & (experience, -0.1) & - \\ 
\midrule
\multirow{5}{*}{FAIR-SD} & (nursing, 0.1) & (\textbf{mrs.}, 0.2) & (\textbf{he}, -0.1) & (mental, -0.2) &(surgeon, 0.3) & (surgery, 0.3) & (group, -0.1) & (\textbf{she}, -0.1) \\ 
& - & (\textbf{she}, 0.1) & - & (research, -0.1) &(surgery, 0.3) & (practice, 0.3) & (general, -0.1) & - \\ 
& - & (nursing, 0.1) & - & (dr., -0.1) &(surgeons, 0.2) & (âģļs, 0.1) & (supports, -0.1) & - \\ 
& - & (nurses, 0.1) & - & (via, -0.1) &(surgical, 0.2) & - & (health, -0.1) & - \\ 
& - & \textbf{(ms.}, 0.1) & - & (who, -0.1) &(surgeries, 0.2) & - & (clinic, -0.1) & - \\ 
\midrule
\multirow{5}{*}{FAIR-DFL} & - & (\textbf{she}, 0.2) & (\textbf{he}, -0.2) & (doctors, -0.1) &(surgeon, 0.6) & (surgery, 0.4) & (each, -0.2) & (\textbf{she}, -0.2) \\ 
& - & (mrs., 0.1) & (medical, -0.1) & (:, -0.1) &(neurosurgery, 0.5) & (shield, 0.1) & (working, -0.1) & (\textbf{her}, -0.1) \\ 
& - & (\textbf{ms.}, 0.1) & - & (other, -0.1) &(surgery, 0.3) & (dr., 0.1) & (care, -0.1) & - \\ 
& - & (\textbf{her}, 0.1) & - & (groups, -0.1) &(surgeons, 0.2) & - & (general, -0.1) & - \\ 
& - & - & - & (., -0.1) &(surgical, 0.2) & - & ((, -0.1) & - \\ 
         \bottomrule
    \end{tabular}
    }
    \caption{Top-attributed positive and negative words based on normalized LRP scores for the unbalanced (biased) version of BIOS. We normalize positive and negative independently using the softmax function and aggregate across all test examples.}
    \label{tab:words}
\end{table*}

\end{document}